\pdfoutput=1

\documentclass[11pt]{article}

\usepackage[preprint]{acl}

\usepackage{times}
\usepackage{latexsym}

\usepackage[T1]{fontenc}

\usepackage[utf8]{inputenc}
\usepackage{microtype}

\usepackage{inconsolata}

\usepackage{graphicx}
\usepackage{booktabs}
\usepackage{enumitem}
\usepackage{lipsum}
\usepackage{float}
\usepackage{tabularx}
\usepackage{soul}
\usepackage{threeparttable}
\usepackage{verbatim}
\usepackage{adjustbox}

\usepackage{xspace}
\newcommand{\method}{LumberChunker\xspace}
\newcommand{\dataset}{GutenQA\xspace}

\title{\method: Long-Form Narrative Document Segmentation}

\author{André V. Duarte\textsuperscript{1}, João Marques\textsuperscript{1}, Miguel Graça\textsuperscript{1},\\ {\bf Miguel Freire\textsuperscript{2}}, {\bf Lei Li\textsuperscript{3}}, {\bf Arlindo Oliveira\textsuperscript{1}}\\
         \textsuperscript{1}INESC-ID / Instituto Superior Técnico, \textsuperscript{2}NeuralShift AI, \textsuperscript{3}Carnegie Mellon University\\
         \texttt{\{andre.v.duarte, joao.p.d.s.marques, arlindo.oliveira\}@tecnico.ulisboa.pt}\\
         \texttt{miguel.graca@inesc-id.pt}\\\
         \texttt{miguel@neuralshift.ai}\\\
         \texttt{leili@cs.cmu.edu}}

\begin{document}
\maketitle
\begin{abstract}
Modern NLP tasks increasingly rely on dense retrieval methods to access up-to-date and relevant contextual information. We are motivated by the premise that retrieval benefits from segments that can vary in size such that a content’s semantic independence is better captured. We propose \method, a method leveraging an LLM to dynamically segment documents, which iteratively prompts the LLM to identify the point within a group of sequential passages where the content begins to shift. To evaluate our method, we introduce \dataset, a benchmark with 3000 ``needle in a haystack'' type of question-answer pairs derived from 100 public domain narrative books available on Project Gutenberg\footnote{Code and Data available at: \url{https://github.com/joaodsmarques/LumberChunker}}. Our experiments show that \method not only outperforms the most competitive baseline by 7.37\% in retrieval performance (DCG@20) but also that, when integrated into a RAG pipeline, \method proves to be more effective than other chunking methods and competitive baselines, such as the Gemini 1.5M Pro.
\end{abstract}

\begin{figure*}[h]
  \centering
  \includegraphics[width=0.9\textwidth]{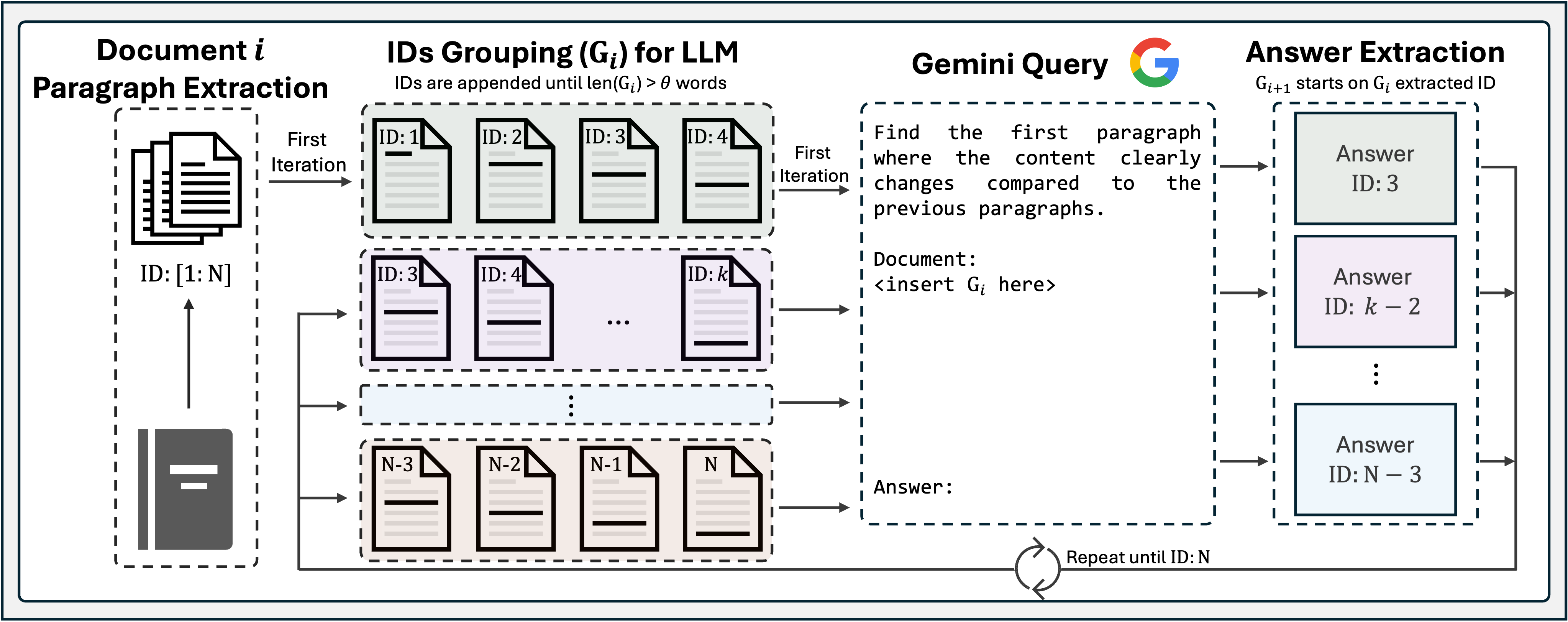}
  \caption{\method follows a three-step process. First, we segment a document paragraph-wise. Secondly, a group ($G_i$) is created by appending sequential chunks until exceeding a predefined token count $\theta$. Finally, $G_i$ is fed as context to Gemini, which determines the ID where a significant content shift starts to appear, thus defining the start of $G_{i+1}$ and the end of the current chunk. This process is cyclically repeated for the entire document.}
    \label{fig:LumberChunker-Pipeline}
  \label{fig:method_full_pipeline}
\end{figure*}

\section{Introduction}

The rapid expansion of Large Language Models (LLMs) has paved the way for tackling a wide range of tasks, including translation \cite{tower}, summarization \cite{summarization}, and question answering \cite{palm}, among others \cite{xuandong_llms}. However, a significant issue arises when these models are tasked with generating responses based on information they lack, often resulting in "hallucinations" - responses that, while seemingly plausible, are factually incorrect \cite{zhang2023siren}. Given the broad accessibility of these models, less informed users may accept all generated content as accurate, potentially causing severe misinterpretations and adverse consequences, like the recent incident where a lawyer cited fictitious cases produced by ChatGPT \cite{ChatGPT} in court, resulting in sanctions for the lawyer and highlighting the severe risks of unverified AI-generated information \cite{law_hallucination}.

\par

In the field of question answering, where precision and accuracy of information are paramount, Retrieval Augmented Generation (RAG) systems present a viable solution to hallucinations by grounding the model's generation on contextually relevant documents \cite{NLP_Rag_paper}.

\par
One often overlooked part of the RAG pipeline is how textual content is segmented into `chunks', which can significantly impact the dense retrieval quality \cite{llms_distracted_by_irrelevant_content}. Real-world applications tend to simplify this step and consider sentences, paragraphs, or propositions as the usual granularity level \cite{passage_strategies_old,proposition_paper}.
\par
In this paper, we propose \method, a novel text segmentation method based on the principle that retrieval efficiency improves when content chunks are as independent as possible from one another. This independence is best achieved by allowing chunks to be of dynamic sizes. Given that language models excel at analyzing text, we leverage their capabilities to identify optimal segmentation points. Specifically, we repeatedly instruct a language model to receive a series of continuous paragraphs and determine the precise paragraph within the sequence where the content starts diverging. This approach ensures that each segment is contextually coherent yet distinct from adjacent segments, thereby enhancing the effectiveness of information retrieval.
\par
We introduce a new benchmark: \dataset, which consists of 100 public domain narrative books manually extracted from Project Gutenberg\footnote{\url{https://www.gutenberg.org/}}. We create 3000 high-quality question-answer pairs from these books to evaluate the impact of \method on retrieval. Finally, we integrate \method into a RAG pipeline for a downstream QA task to assess its effect on the accuracy of the generated outputs.

\section{Background}

The retrieval granularity at which a document is segmented plays an essential role as ineffective chunking strategies can lead to chunks with incomplete context or excessive irrelevant information, which damage the performance of retriever models \cite{llms_distracted_chain_of_note}.
\par
Beyond typical granularity levels like sentences or paragraphs \cite{RAG_survey}, other advanced methods can be employed. Recursive character splitting \cite{recursive} segments text based on a hierarchy of separators such as paragraph breaks, new lines, spaces, and individual characters. While this method better respects the document’s structure, it may lack contextual understanding. To address this, semantic-based splitting \cite{semantic_chunking} utilizes embeddings to cluster semantically similar text segments. This approach ensures that chunks maintain meaningful context and coherence by identifying breakpoints based on significant changes in embedding distances.
\par
The recent research by \citet{proposition_paper} introduces a novel retrieval granularity termed \textit{propositions} - minimal textual units, each conveying an individual fact in a clear, self-sufficient natural language format. While this concept is valid for contexts with fact-based information, like Wikipedia, it may be less effective for narrative texts where the flow and contextual continuity play a critical role (as illustrated in Appendix \ref{sec:propositions_example}).
\par
Retrieval granularity is often viewed at the document level, but it can also involve adjusting the query itself. \citet{hyde_paper} suggests the Hypothetical Document Embeddings (HyDE) method, where an LLM transforms the query into a potential answer document.

\section{Methodology}

\subsection{\method}

Our main proposed contribution is a novel method for document segmentation named \method, which employs an LLM to dynamically segment documents into semantically independent chunks. Our approach is grounded in the principle that retrieval benefits from segments that can vary in size to better capture the content's semantic independence. This dynamic granularity ensures that each chunk encapsulates a complete, standalone idea, enhancing the relevance and clarity of the retrieved documents. By feeding the LLM a set of sequential passages, \method autonomously determines the most appropriate points for segmentation. This decision process takes into account the structure and semantics of the text, thereby enabling the creation of chunks that are optimally sized and contextually coherent.
\par
Figure \ref{fig:LumberChunker-Pipeline} displays the overall pipeline of \method. We start by splitting the target document paragraph-wise, with each paragraph being uniquely identified by an incremental ID number. Each paragraph is sequentially concatenated into a group $G_i$ until its collective token count surpasses a pre-determined threshold, \( \theta \), which is strategically set based on empirical insights, further discussed in \ref{sec:tuning_theta}. The goal is to set $\theta$ large enough to avoid bisecting relevant larger segments while ensuring it is small enough to prevent overwhelming the model with excessive context, which could hinder its reasoning accuracy. The group $G_i$ is given as input to the LLM (we choose Gemini 1.0-Pro \cite{gemini_1.0}), which we instruct to pinpoint the specific paragraph within \( G_i \) where the content is perceived to diverge significantly from the preceding context. This detection marks the end of a chunk. The document keeps being sequentially partitioned into chunks in a cyclical manner, with the starting point of each new \( G_{i+1} \) group being the paragraph identified in the previous iteration. The prompt used is provided in Appendix \ref{sec:lumberchunker_prompt}.

\subsection{GutenQA}

Our proposed benchmark comprises a collection of 100 books sourced from Project Gutenberg. Due to the diverse HTML structures of these books, we extract the content manually. This avoids potential errors associated with automatic text extraction, as discussed in Appendix \ref{sec:manual_extraction}.
\par
We use ChatGPT (\texttt{gpt-3.5-turbo-0125}) to generate questions for each book. Initially, more than 10000 questions are automatically generated, which are then filtered down such that each book has 30 high-quality questions. To evaluate the method's retrieval capabilities, we specifically design questions to be factual and specific, targeting information unlikely to be repeated elsewhere in the text. This selection strategy favors `what,' `when,' and `where' questions over `why' and `how' questions. The prompt used to instruct the model to generate questions, along with statistics about the distribution of question types within the dataset, is provided in Appendix \ref{sec:appendix_test_data_generation}.

\begin{table*}[h]
\centering
\caption{Passage retrieval performance (DCG@$k$ and Recall@$k$) on \dataset with different granularities on the questions\textsuperscript{\textdagger} and on the retrieval corpus passages. The best scores in each column are highlighted in \textbf{bold}.}
\label{tab:retrieval_results}
\begin{threeparttable}
\begin{adjustbox}{max width=\textwidth}
\begin{tabular}{@{}lccccccccccc@{}}
\toprule
\multicolumn{1}{c}{} & \multicolumn{5}{c}{\textbf{DCG @ $k$}} &  & \multicolumn{5}{c}{\textbf{Recall @ $k$}} \\ \cmidrule(lr){2-6} \cmidrule(l){8-12} 
\multicolumn{1}{c}{} & 1 & 2 & 5 & 10 & 20 &  & 1 & 2 & 5 & 10 & 20 \\ \cmidrule(r){1-6} \cmidrule(l){8-12} 
Semantic Chunking & 29.50 & 35.31 & 40.67 & 43.14 & 44.74 &  & 29.50 & 38.70 & 50.60 & 58.21 & 64.51 \\
Paragraph-Level & 36.54 & 42.11 & 45.87 & 47.72 & 49.00 &  & 36.54 & 45.37 & 53.67 & 59.34 & 64.34 \\
Recursive Chunking & 39.04 & 45.37 & 50.66 & 53.25 & 54.72 &  & 39.04 & 49.07 & 60.64 & 68.62 & 74.35 \\
HyDE\textsuperscript{\textdagger} & 33.47 & 39.74 & 45.06 & 48.14 & 49.92 &  & 33.47 & 43.41 & 55.11 & 64.61 & 71.61\\
Proposition-Level & 36.91 & 42.42 & 44.88 & 45.65 & 46.19 &  & 36.91 & 45.64 & 51.04 & 53.41 & 55.54 \\
\method & \textbf{48.28} & \textbf{54.86} & \textbf{59.37} & \textbf{60.99} & \textbf{62.09} &  & \textbf{48.28} & \textbf{58.71} & \textbf{68.58} & \textbf{73.58} & \textbf{77.92} \\ 
\bottomrule
\end{tabular}
\end{adjustbox}
\end{threeparttable}
\end{table*}

\section{Experiments}

We evaluate \method using a series of diverse experiments. The key questions that guide our experimental evaluation are as follows:
\begin{itemize}[label=•, leftmargin=*]

\item \textbf{What is the optimal threshold for target token count in each \method prompt?} We analyze how LumberChunker's DCG@$k$ and Recall@$k$ scores are influenced by different prompt lengths $\theta \in [450, 1000]$ tokens.

\item \textbf{Does \method enhance retrieval?} We evaluate \method’s ability to locate highly specific information within the documents, as represented by our \dataset questions. We compare its DCG@$k$ and Recall@$k$ scores against other baseline methods, such as Semantic or Proposition-Level chunking.

\item \textbf{Do \method chunks enhance generation quality?} It is natural to question whether the increased computational cost of segmenting a document with our method is worthwhile. To address this, we evaluate if \method chunks improve generation quality in a QA task. For this purpose, we integrate our chunks into a RAG pipeline and create a smaller QA test set comprising 280 questions based on four narrative autobiographies. As these elements are outside the paper's main scope, further details are provided in Appendix \ref{sec:appendix_rag_pipeline_autobiographies}. We compare our approach with other RAG pipeline variants using different chunking techniques, including manually created chunks, which we consider the gold standard for optimal retrieval. We also employ non-RAG baselines like Gemini 1.5 Pro \cite{gemini_1.5}, capable of processing up to 1.5 million input tokens, and a closed-book setup where Gemini Pro relies solely on internal knowledge.

\end{itemize}

\section{Results and Discussion}

\subsection{Context Size}
\label{sec:tuning_theta}

\begin{figure}
    \centering
    \includegraphics[width=1\linewidth]{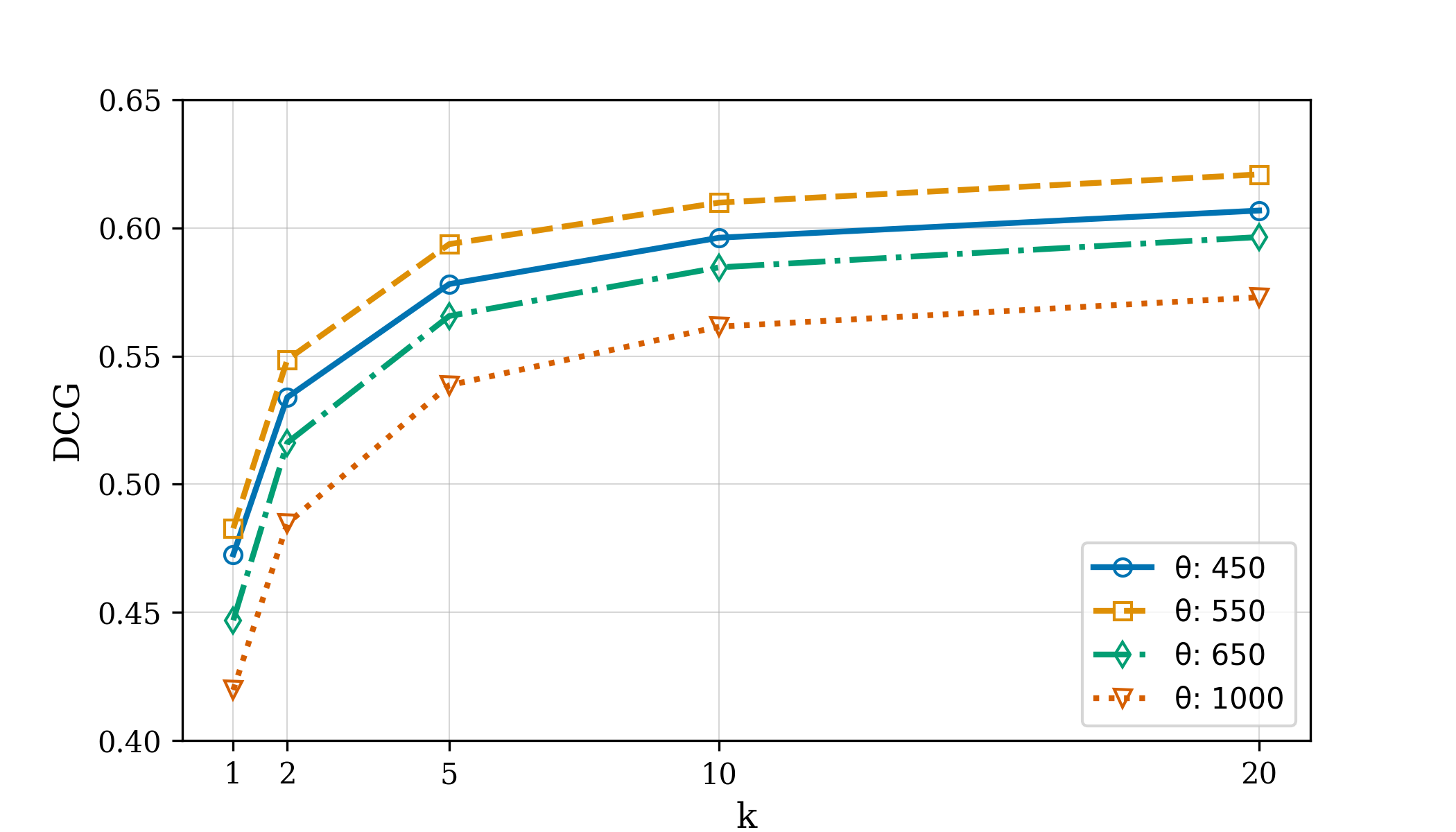}
    \caption{Optimizing Context Size $\theta$ ($\approx$ number of tokens in the \method prompt.)}
    \label{fig:dcg_theta_tune}
\end{figure}

Figure \ref{fig:dcg_theta_tune} reveals that among the various thresholds tested, \(\theta = 550\) leads to the best performance, achieving the highest DCG@$k$ scores across all values of \(k\) tested. This indicates that prompts with around 550 tokens optimize the quality of document retrieval by effectively balancing context capture and passage length. Following this, thresholds \(\theta = 450\) and \(\theta = 650\) show similar but slightly lower performances, suggesting that while they are effective, they do not capture the optimal balance as well as \(\theta = 550\). The threshold \(\theta = 1000\) performs the worst, with noticeably lower DCG@$k$ scores. Given that this task requires advanced reasoning, an excessively long prompt may overwhelm the model's capacity to focus on the relevant parts of the input, thus compromising its performance.

\subsection{Main Results}

The results presented in Table \ref{tab:retrieval_results} highlight that for all values of \(k\), \method consistently outperforms every other baseline both on DCG@$k$ and Recall@$k$ metrics\footnote{Chunks are encoded with \texttt{text-embedding-ada-002} embeddings from OpenAI.}. This is particularly evident at \(k = 20\), where LumberChunker's DCG score reaches 62.09, while the closest competitor, Recursive chunking, only achieves a score of 54.72. Similarly, in terms of Recall@$k$, \method attains a score of 77.92 at \(k = 20\), compared to Recursive Chunking's 74.35.
\par
A closer examination of the baselines reveals that methods like Paragraph-Level and Semantic chunking fail to scale effectively as $k$ increases, indicating their limitations in maintaining relevance over a larger number of retrieved documents. HyDE, which uses Recursive chunking as its document granularity level, also fails to outperform its simpler counterpart for every value of $k$. This suggests that the additional augmentation layer the HyDE introduces may not be suited for this task.
\par
The scores for Proposition-Level chunking are notably lower than those of \method. While Proposition-Level chunking excels in contexts with fine-grained, fact-based information, such as Wikipedia text, it is less effective for narrative texts where flow and contextual continuity play a critical role. For details on the segmentation of \dataset, refer to Appendix \ref{sec:granularity_statistics}.

\subsection{Impact on QA Systems}

\begin{figure}[h]
    \centering
    \includegraphics[width=1\linewidth]{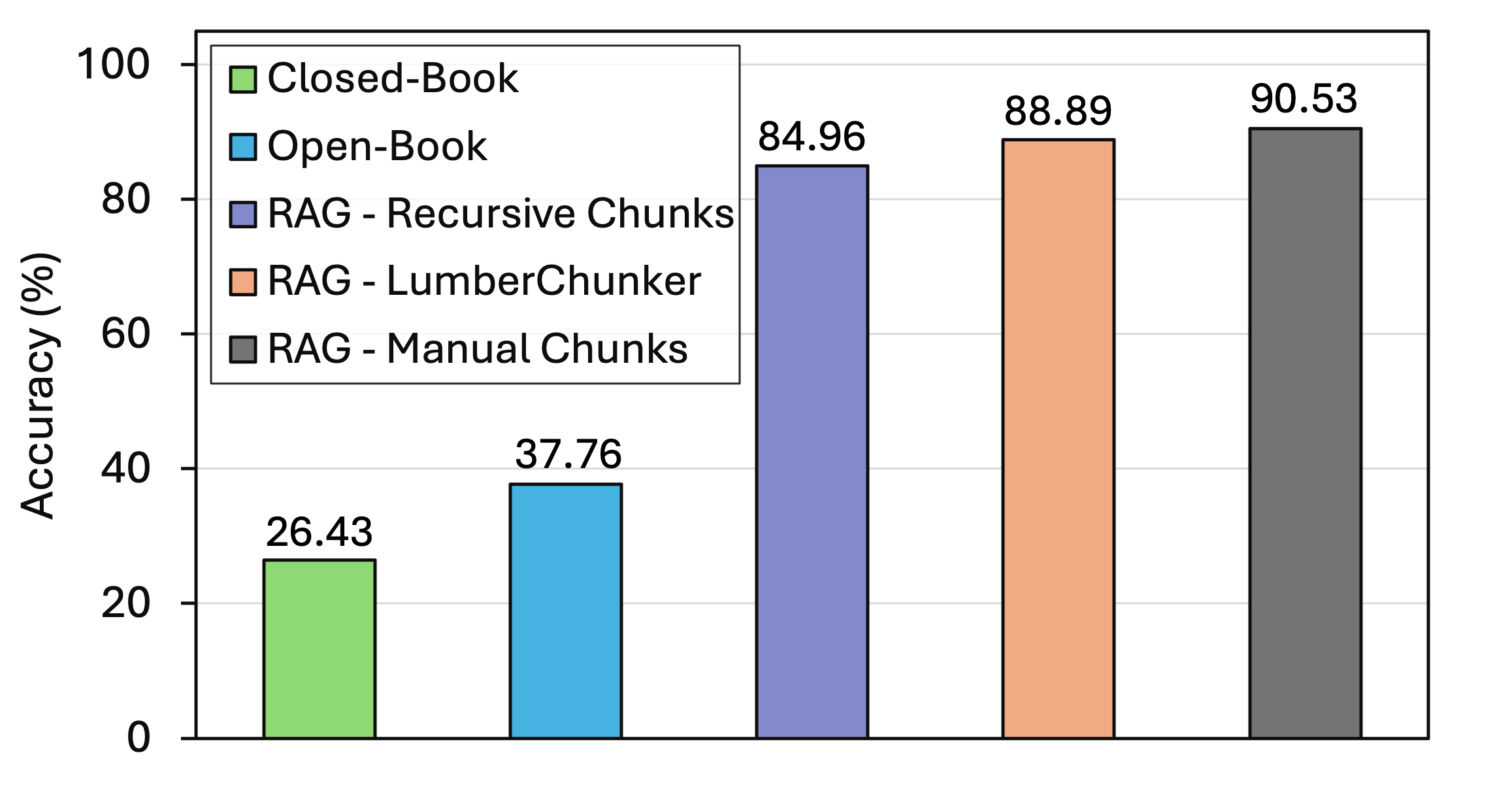}
    \caption{QA Accuracy on Autobiographies Test Set.}
    \label{fig:rag_accuracy}
\end{figure}

In Figure \ref{fig:rag_accuracy}, we observe the performance of different QA methodologies applied to several autobiographies. Notably, our proposed \method, when integrated into the RAG pipeline, demonstrates superior average accuracy compared to most baselines. Particularly, it outperforms again Recursive chunking, which was the most competitive baseline in the retrieval evaluation (Table \ref{tab:retrieval_results}). This reinforces our view that retrieval has a positive impact on accuracy. \method only falls short of the RAG approach with manual chunks, which is the expected gold standard in this task.

\section{Conclusions}

In this study, we introduce \method, a novel text segmentation method leveraging LLM textual capabilities to dynamically segment documents into semantically coherent chunks.
\par
We also present \dataset, a collection of 100 carefully parsed public domain books, segmented with \method, for which we create 3000 curated question-answer pairs.

\section{Limitations}
Despite \method demonstrating superior performance compared to all baselines, it should be highlighted that it requires the use of an LLM, which automatically renders it more expensive and slower compared to traditional methods like Recursive chunking (further discussed in Appendix \ref{sec:computational_efficiency}).
\par
\method is designed for narrative texts, which are somewhat unstructured and benefit from semantic textual interpretation. However, for scenarios with highly structured texts like those in the legal domain, \method may be an unnecessarily complex solution because it would likely achieve similar segmentations as those employing document-wise structure parsing (like Markdown segmentation), but at a more expensive cost.
\par
Within the present methodology, \method also faces scalability issues with the length of individual documents and the volume of documents that need to be processed. While each document needs to be processed only once, the iterative nature of prompting the language model to identify segmentation points can become a drawback when dealing with large number of documents.

\section{Ethical Considerations}

This paper focuses on improving text segmentation methods by leveraging existing large language models. Our released dataset, GutenQA, uses only public domain texts, ensuring there are no privacy concerns or handling of sensitive data. Regarding LumberChunker, we do not foresee any direct impact on malicious activities, disinformation, surveillance, or any significant environmental impact beyond the typical computational requirements.
\par
The only ethical consideration we would like to highlight is our extensive use of black box models. Unlike traditional chunking techniques like Recursive Chunking, which are fully transparent and easily reproducible, black box models introduce some uncertainty regarding their outputs. As a result, it is not impossible that our methodology might have some biases we are unaware of.

\appendix
\onecolumn

\section{Propositions Example on Narrative Texts}
\label{sec:propositions_example}

\begin{table}[H]
\setlength\extrarowheight{6pt}
  \centering
    \caption{Example of \textbf{\textcolor{ForestGreen}{good}} passage transformation into propositions.}

    \label{tab:good_proposition}
  \begin{tabularx}{\textwidth}{X}
    \toprule[1.1pt] 
    \textbf{Passage 1:} \textit{Elon Musk’s attraction to risk was a family trait. In that regard, he took after his maternal grandfather, \textbf{Joshua Haldeman}, a daredevil adventurer with strongly held opinions who was raised on a farm on the barren plains of central Canada. \textbf{He} studied chiropractic techniques in Iowa, then returned to his hometown near Moose Jaw, where he broke in horses and gave chiropractic adjustments in exchange for food and lodging.}
\par \
\par
\textbf{Proposition 1.1}: Elon Musk’s attraction to risk was a family trait.\\
\textbf{Proposition 1.2}: His maternal grandfather was Joshua Haldeman.\\
\textbf{Proposition 1.3}: Joshua Haldeman was a daredevil adventurer with strongly held opinions.\\
\textbf{Proposition 1.4}: Joshua Haldeman was raised on a farm on the barren plains of central Canada.\\
\textbf{Proposition 1.5}: Joshua Haldeman studied chiropractic techniques in Iowa.\\
\textbf{Proposition 1.6}: Joshua Haldeman returned to his hometown near Moose Jaw.\\
\textbf{Proposition 1.7}: In his hometown, Joshua Haldeman broke in horses and gave chiropractic adjustments.\\
\textbf{Proposition 1.8}: Joshua Haldeman received food and lodging in exchange for his chiropractic services.\\

\bottomrule[1.1pt] 
\end{tabularx}
\end{table}

\begin{table}[H]
\setlength\extrarowheight{6pt}
  \centering
    \caption{Example of \textbf{\textcolor{BrickRed}{poor}} passage transforming into propositions.}
    \label{tab:bad_proposition}
  \begin{tabularx}{\textwidth}{X}
    \toprule[1.1pt] 
    \textbf{Passage 2:} \textit{\textbf{He} was eventually able to buy his own farm, but he lost it during the depression of the 1930s. For the next few years, he worked as a cowboy, rodeo performer, and construction hand. His one constant was a love for adventure. \textbf{He} married and divorced, traveled as a hobo on freight trains, and was a stowaway on an oceangoing ship.}
\par \
\par
\textbf{Proposition 2.1}: He was eventually able to buy his own farm.\\
\textbf{Proposition 2.2}: He lost his own farm during the depression of the 1930s.\\
\textbf{Proposition 2.3}: For the next few years, he worked as a cowboy.\\
\textbf{Proposition 2.4}: For the next few years, he worked as a rodeo performer.\\
\textbf{Proposition 2.5}: For the next few years, he worked as a construction hand.\\
\textbf{Proposition 2.6}: His one constant was a love for adventure.\\
\textbf{Proposition 2.7}: He married and divorced.\\
\textbf{Proposition 2.8}: He traveled as a hobo on freight trains.\\
\textbf{Proposition 2.9}: He was a stowaway on an oceangoing ship.\\
\bottomrule[1.1pt] 
\end{tabularx}
\end{table}

\noindent \textbf{Comment:} Unlike the example in Table \ref{tab:good_proposition}, the pronoun `\textbf{He}' in Table \ref{tab:bad_proposition} passage cannot be accurately co-referenced, resulting in somewhat ambiguous propositions. Consequently, if a user asks a question like 'Who in Elon Musk's family worked as a rodeo performer?', a model that uses only propositions as retrieval units will not be able to provide an accurate response.\\

\newpage

\section{\method Gemini Prompt}
\label{sec:lumberchunker_prompt}

\begin{table}[H]
  \centering
    \caption{\method Gemini Prompt example for the book: \textit{Winnie the Pooh} by A. A. Milne.}
  \begin{tabularx}{\textwidth}{X}
    \toprule[1.1pt] 
    \textbf{Prompt:} You will receive as input an English document with paragraphs identified by `ID XXXX: <text>'.
\par \ 
\par
\textbf{Task:} Find the first paragraph (not the first one) where the content clearly changes compared to the previous paragraphs.

\par \ 
\par
\textbf{Output:} Return the ID of the paragraph with the content shift as in the exemplified format: `Answer: ID XXXX'.
\par \ 
\par
\textbf{Additional Considerations:} Avoid very long groups of paragraphs. Aim for a good balance between identifying content shifts and keeping groups manageable.
\par \
\par
\textbf{Document:}
\par
ID 0001: \textit{Here is Edward Bear, coming downstairs now, bump, bump, bump, on the back of his head, behind Christopher Robin. It is, as far as he knows, the only way of coming downstairs, but sometimes he feels that there really is another way, if only he could stop bumping for a moment and think of it. And then he feels that perhaps there isn't. Anyhow, here he is at the bottom, and ready to be introduced to you. Winnie-the-Pooh.}
\par \ 
\par
ID 0002: \textit{When I first heard his name, I said, just as you are going to say, "But I thought he was a boy?"}
\par \ 
\par
ID 0003: \textit{"So did I," said Christopher Robin.}
\par \ 
\par
ID 0004: \textit{"Then you can't call him Winnie?"}
\par \ 
\par
ID 0005: \textit{"I don't."}
\par \
\par \
...
\par \
\par \
ID 0018: \textit{So I tried.}
\par \
\par \
ID 0019: \textit{Once upon a time, a very long time ago now, about last Friday, Winnie-the-Pooh lived in a forest all by himself under the name of Sanders.}
\par \
\par \
ID 0020: \textit{"What does 'under the name' mean?" asked Christopher Robin.}
\par \
\par \
\par \
\par \
\textbf{Answer:} ID 0019\\
\bottomrule[1.1pt] 
\end{tabularx}
\end{table}

\newpage

\section{Project Gutenberg - Manual Extraction}
\label{sec:manual_extraction}

\begin{table}[H]
  \centering
    \caption{Passage manually extracted from Project Gutenberg regarding the book: \textit{Anna Karenina}, by Leo Tolstoy.}
       \label{tab:parsing_gutenqa}
  \begin{tabularx}{0.95\textwidth}{X}
    \toprule[1.1pt] 
    \textbf{\dataset Verbatim:} They were carrying something, and dropped it.
\par “I told you not to sit passengers on the roof,” said the little girl in English; “there, pick them up!”
\par “Everything’s in confusion,” thought Stepan Arkadyevitch; “there are the children running about by themselves.” And going to the door, he called them. They threw down the box, that represented a train, and came in to their father.
\par The little girl, her father’s favorite, ran up boldly, embraced him, and hung laughingly on his neck, enjoying as she always did the smell of scent that came from his whiskers. At last the little girl kissed his face, which was flushed from his stooping posture and beaming with tenderness, loosed her hands, and was about to run away again; but her father held her back.
\par “How is mamma?” he asked, passing his hand over his daughter’s smooth, soft little neck. “Good morning,” he said, smiling to the boy, who had come up to greet him. He was conscious that he loved the boy less, and always tried to be fair; but the boy felt it, and did not respond with a smile to his father’s chilly smile.\\              
    \bottomrule[1.1pt]
  \end{tabularx}
\end{table}

\begin{table}[H]
  \centering
    \caption{Passage from NarrativeQA \cite{narrativeqa} regarding the book: \textit{Anna Karenina}, by Leo Tolstoy.}
    \label{tab:parsing_narrativeqa}
  \begin{tabularx}{0.95\textwidth}{X}
    \toprule[1.1pt] 
    \textbf{NarrativeQA Verbatim:} They were carrying something, and dropped it.\textbackslash n\textbackslash nâ\textbackslash x80\textbackslash x9cI told you not to sit passengers on the roof,â\textbackslash x80\textbackslash x9d said the little girl in\textbackslash nEnglish; â\textbackslash x80\textbackslash x9cthere, pick them up!â\textbackslash x80\textbackslash x9d\textbackslash n\textbackslash nâ\textbackslash x80\textbackslash x9cEverythingâ\textbackslash x80\textbackslash x99s in confusion,â\textbackslash x80\textbackslash x9d thought Stepan Arkadyevitch; â\textbackslash x80\textbackslash x9cthere are\textbackslash nthe children running about by themselves.â\textbackslash x80\textbackslash x9d And going to the door, he\textbackslash ncalled them. They threw down the box, that represented a train, and\textbackslash ncame in to their father.\textbackslash n\textbackslash nThe little girl, her fatherâ\textbackslash x80\textbackslash x99s favorite, ran up boldly, embraced him,\textbackslash nand hung laughingly on his neck, enjoying as she always did the smell\textbackslash nof scent that came from his whiskers. At last the little girl kissed\textbackslash nhis face, which was flushed from his stooping posture and beaming with\textbackslash ntenderness, loosed her hands, and was about to run away again; but her\textbackslash nfather held her back.\textbackslash n\textbackslash nâ\textbackslash x80\textbackslash x9cHow is mamma?â\textbackslash x80\textbackslash x9d he asked, passing his hand over his daughterâ\textbackslash x80\textbackslash x99s smooth,\textbackslash nsoft little neck. â\textbackslash x80\textbackslash x9cGood morning,â\textbackslash x80\textbackslash x9d he said, smiling to the boy, who had\textbackslash ncome up to greet him. He was conscious that he loved the boy less, and\textbackslash nalways tried to be fair; but the boy felt it, and did not respond with\textbackslash na smile to his fatherâ\textbackslash x80\textbackslash x99s chilly smile.\\       
    \bottomrule[1.1pt]
  \end{tabularx}
\end{table}

\noindent \textbf{Comment:} The differences between the passages on Table \ref{tab:parsing_gutenqa} and Table \ref{tab:parsing_narrativeqa} are significant. The latter passage is marked by misplaced `\textbackslash n' characters and the presence of non-standard characters such as `â' and `\textbackslash x80\textbackslash x9c', which likely result from encoding issues during the data extraction process. In contrast, the first passage, manually extracted from Project Gutenberg, is more readable and free from such errors.

\newpage

\section{Artificial Test Data Generation - \dataset}
\label{sec:appendix_test_data_generation}

\begin{table}[H]
  \centering
    \caption{Test set questions prompt template for the book: \textit{Anna Karenina}, by Leo Tolstoy.}
  \begin{tabularx}{0.95\textwidth}{X}
    \toprule[1.1pt] 
    \textbf{System Prompt:} Your task is to generate a question-answer pair that is specific to the provided text excerpts from the book "Anna Karenina" by Leo Tolstoy. The question should be unique to the passage, meaning it cannot be easily answered by other parts of the book.
\par
\textbf{Instructions:}
\par \ 
\par
\textbf{Read the Passage:} Carefully read the provided text excerpt from the book. Understand the context, key events, and specific details mentioned.
\par \ 
\par
\textbf{Formulate a Question:} Create a question that is:
\par \ 
\par
\textbf{Directly related to the passage:} The question should be based on the specific information or events described in the text.
\par \ 
\par
\textbf{Unique to the passage:} The question should not be answerable with information from other parts of the book.
\par \ 
\par
\textbf{Type:} Focus on creating a "When/What/Where" question to encourage specificity and conciseness.
\par \ 
\par
\textbf{Provide a Concise Answer:} Write an answer that is:
\par \ 
\par
\textbf{Direct and informative:} Limit the answer to a maximum of two sentences. Ensure it directly addresses the question and is supported by the passage.
\par \ 
\par
\textbf{Self-contained:} The answer should make sense on its own and should not require additional context from outside the passage.
\par \ 
\par
\textbf{Cite the Supporting Passage:} Include the passage that contains the information needed to answer the question. This will be used to verify the accuracy of the answer and the relevance of the question. Do not use `...'. The passage should be quoted without breaks.\\

    \midrule 

    \textbf{User Prompt:} \textit{Example:}
\par
Passage: \textit{“Vous comprenez l’anglais?” asked Lidia Ivanovna, and receiving a reply in the affirmative, she got up and began looking through a shelf of books. “I want to read him ‘Safe and Happy,’ or ‘Under the Wing,’” she said, looking inquiringly at Karenin.”}
\par \ 
\par
Question: \textit{What book does Countess Lidia Ivanovna want to read to Karenin?}
\par \ 
\par
Answer: \textit{She wants to read him 'Safe and Happy,' or 'Under the Wing.'}
\par \ 
\par
Supporting Passage: \textit{“I want to read him ‘Safe and Happy,’ or ‘Under the Wing,’” she said, looking inquiringly at Karenin.}\\              
    \bottomrule[1.1pt]
  \end{tabularx}
\end{table}


\begin{table}[H]
\centering
\caption{Frequency of the first token in \dataset questions.}
\label{tab:token_frequency}
\begin{tabular}{@{}lccccccc@{}}
\toprule
\textbf{} & \textbf{What} & \textbf{How} & \textbf{Why} & \textbf{Who} & \textbf{Where} & \textbf{When} & \textbf{Other} \\ \midrule
\textbf{Frequency} & 49.3\% & 20.1\% & 15.2\% & 7.7\% & 2.8\% & 0.3\% & 4.5\% \\ 
\bottomrule
\end{tabular}
\end{table}

\newpage 

\section{\method impact on Generation - Details}
\label{sec:appendix_rag_pipeline_autobiographies}

\subsection{RAG Pipeline}
\begin{figure}[h]
    \centering
    \includegraphics[width=0.95\linewidth]{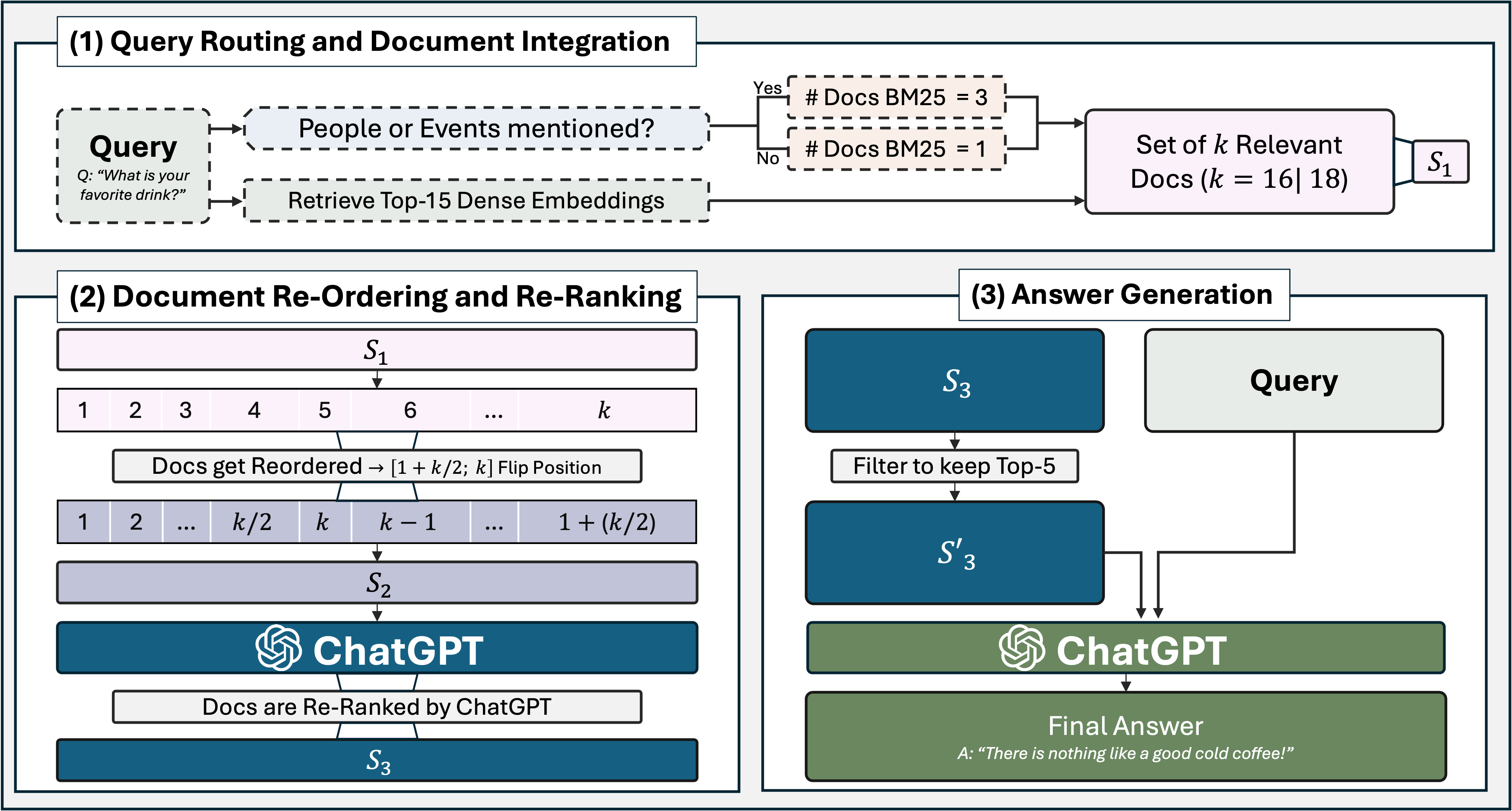}
    \caption{RAG Pipeline for QA on Autobiographies}
    \label{fig:rag_pipeline}
\end{figure}

We build a RAG-based QA pipeline specifically tailored to biographical books. We employ a hybrid retrieval format, combining OpenAI \texttt{text-ada-embedding-002} dense embeddings with BM25. Based on Figure \ref{fig:rag_pipeline}, we outline the step-by-step process employed in our system.
\par
\par
\textbf{Query Routing and Document Integration:} Each query undergoes an evaluation by a detector that identifies mentions of people or events. If the detector identifies relevant mentions within the query, the top-3 most relevant chunks are retrieved using the BM25 algorithm. A single document is retrieved as a precautionary measure if no mentions are detected (acknowledging occasional detector failures). Concurrently, the top-15 chunks are retrieved through a dense retrieval mechanism. This step is designed to enhance the retrieval quality by accessing deeper semantic relations that BM25 might miss. We implement an intersection check between the documents retrieved by BM25 and those from dense retrieval, and overlapping documents from BM25 are removed to avoid redundancy. The highest-ranked document from BM25 is then prioritized at the top of the retrieval list, with the second and third (if available) placed at the end, ensuring a blend of retrieval strategies.
\par
\ 
\par
\textbf{Document Re-Ordering and Re-Ranking:} The retrieved chunks are integrated into the context window of ChatGPT (gpt-3.5-turbo). If the context includes six or more chunks, we employ a strategy to reverse the order of the chunks from the midpoint onwards. This re-ordering aims to potentially minimize the model's `Lost in the Middle' problem, where model performance degrades for information located in the middle of long contexts but starts recovering towards the end, creating a U-shaped performance curve \cite{lost_in_the_middle}. ChatGPT is then prompted to identify and re-order the documents based on their decreasing relevance to the query.
\par
\ 
\par
\textbf{Final Answer Generation:} The response is generated in this final step. The top-5 documents, as determined by the model, are retained for the final answer generation. The model synthesizes the information from the top documents into a coherent and contextually accurate answer, aiming to address the query comprehensively.

\subsection{Autobiography Questions}

We created a dataset based on four autobiographies: (1) Mahatma Gandhi, (2) Helen Keller, (3) Benjamin Franklin, and (4) Francisco Pinto Balsemão \cite{gandhi,keller,franklin,balsemao}. Each book includes 70 questions, a combination of manually crafted and ChatGPT-generated ones. Specifically, 40 questions are auto-generated following the methodology described for the \dataset, while the remaining 30 questions are manually created by a human while reading the book.
\par
Each book contains manually created questions and is also manually segmented, as one of the baselines for the task is a RAG Pipeline with manual chunks. Both tasks are time-consuming but crucial for ensuring the true impact of \method on the generation quality.

\section{\method and Baseline methods - Chunks Statistics}
\label{sec:granularity_statistics}

\begin{table}[h]
\centering
\caption{The average number of tokens on each chunk after segmenting every \dataset book.}
\label{tab:word_counts}
\begin{tabular}{@{}llc@{}}
\toprule
\textbf{} &  & \textbf{Avg. \#Tokens / Chunk} \\ \cmidrule(r){1-1} \cmidrule(l){3-3} 
Semantic Chunking &  & 185 tokens \\
Paragraph Level &  & 79 tokens \\
Recursive Chunking &  & 399 tokens \\
Proposition-Level &  & 12 tokens \\ 
\method &  & 334 tokens \\ 

\bottomrule
\end{tabular}
\end{table}

From Table \ref{tab:word_counts}, we observe that the average number of tokens per chunk for \method is approximately 40\% below the intended input size of 550 tokens. This suggests that, on average, the LLM does not frequently select IDs near the end of the input context. This is a positive sign, as consistently choosing IDs near the end could imply the model is not reasoning over the input effectively, leading to the 'lost in the middle' problem \cite{lost_in_the_middle}.
\par
Both Paragraph, Recursive, and Proposition-Level chunking exhibit values that align with expectations. The high prevalence of dialogue in the dataset explains the relatively small average number of tokens per chunk at the paragraph level. Recursive chunking achieves an expected value since the \texttt{RecursiveCharacterTextSplitter} from \texttt{langchain}\footnote{\url{https://python.langchain.com/v0.1/docs/modules/data_connection/document_transformers/recursive_text_splitter/}} was configured with an optimal value of 450 tokens in mind. Additionally, the average chunk size for propositions is consistent with findings by \citet{proposition_paper}, which reported an average proposition length of 11.2 tokens for the processed Wikipedia corpus.
\par
On the other hand, Semantic chunking appears to have a relatively small average token size. We hypothesize that this is primarily due to the nature of the documents (narrative books), which often contain significant amounts of dialogue, resulting in short paragraphs. Consequently, Semantic chunking may not fully capture the broader semantic context intended for each paragraph, thus fragmenting the text more than necessary.

\newpage

{\section{Computational Efficiency - \method and Baselines}
\label{sec:computational_efficiency}

\begin{table}[H]
\centering
\caption{The time required to apply \method or one of the baselines on each book.}
\label{tab:lumberchunker-time2}
\begin{tabular}{@{}llccc@{}}
\toprule
\textbf{} &  & \multicolumn{3}{c}{\textbf{Avg. Seconds to Complete a Book}} \\ \cmidrule(l){3-5} 
 &  & \begin{tabular}[c]{@{}c@{}}A Christmas Carol\\ (710 Paragraphs)\end{tabular} &  & \begin{tabular}[c]{@{}c@{}}The Count of Monte Cristo\\ (14339 Paragraphs)\end{tabular} \\ \cmidrule(r){1-1} \cmidrule(lr){3-3} \cmidrule(l){5-5} 
Semantic Chunking &  & 212 seconds &  &  4978 seconds \\
Recursive Chunking &  & 0.1 seconds &  &  0.6 seconds \\
HyDE &  & 75 seconds &  &  79 seconds \\
Proposition-Level &  & 633 seconds &  &  10302 seconds \\
\method &  & 95 seconds &  &  1628 seconds \\ \bottomrule
\end{tabular}
\end{table}

From Table \ref{tab:lumberchunker-time2}, we observe the impact of document size on the completion times of the tested approaches.  Recursive chunking, although demonstrating a sixfold time increase between both books, is still the faster method. We attribute this efficiency mainly to the fact that Recursive chunking does not involve any LLM API requests. HyDE, despite employing an LLM, maintains a constant number of LLM queries per book (always 30 API requests), resulting in its completion time being invariant to the document size. On the other hand, \method, Semantic, and Proposition-Level chunking exhibit significant increases in completion time with larger documents. It is important to note that both Semantic and Proposition-Level chunking can be optimized by making asynchronous OpenAI API requests, significantly reducing completion times. However, \method does not allow for such optimization because its methodology requires dynamic queries to the LLM that cannot be pre-established. While \method does enhance retrieval performance over every other baseline, we acknowledge the potential for further optimization.
}

\end{document}